\pdfoutput=1

\documentclass[11pt]{article}

\usepackage[final]{acl}

\usepackage{times, soul}
\usepackage{latexsym}

\usepackage[T1]{fontenc}

\usepackage[utf8]{inputenc}

\usepackage{microtype}

\usepackage{inconsolata}

\usepackage{graphicx}

\usepackage{enumitem}
\usepackage{multirow}
\usepackage{multicol}
\usepackage{colortbl}
\usepackage{bbding}
\usepackage{pifont}
\usepackage{wasysym}
\usepackage{amssymb}
\usepackage[normalem]{ulem}
\useunder{\uline}{\ul}{}
\usepackage{float}
\usepackage[flushleft]{threeparttable}
\usepackage{booktabs}
\usepackage[skip=0pt]{caption}
\usepackage{subcaption}
\usepackage{makecell}
\usepackage[symbol]{footmisc}

\newcommand{\method}[1]{\textsc{#1}}
\newcommand{\Ours}{\method{EDF}}
\newcommand{\OursLong}{Entity Decomposition with Filtering}

\newcommand{\sY}{\mathcal{Y}}

\newcommand{\sD}{\mathcal{D}}

\newcommand{\sS}{\mathcal{S}}

\newcommand{\bx}{\mathbf{x}}

\newcommand{\bs}{\mathbf{s}}

\newcommand{\bt}{\mathbf{t}}
\newcommand{\shY}{\hat{\mathcal{Y}}}
\newcommand{\cmark}{\ding{51}}%
\newcommand{\xmark}{\ding{55}}%

\begin{document}

%
%

\title{Entity Decomposition with Filtering:\\ A Zero-Shot Clinical Named Entity Recognition Framework}



\author{Reza Averly\textsuperscript{$\clubsuit$} \ Xia Ning\textsuperscript{$\clubsuit$}\textsuperscript{$\spadesuit$}\textsuperscript{$\heartsuit$}\textsuperscript{$\diamondsuit$} \\
\textsuperscript{$\clubsuit$} Department of Computer Science
and Engineering, The Ohio State University, USA\\
\textsuperscript{$\spadesuit$}Department of Biomedical Informatics, The Ohio State University, USA\\
\textsuperscript{$\heartsuit$}Translational Data Analytics Institute, The Ohio State University, USA\\
\textsuperscript{$\diamondsuit$}College of Pharmacy, The Ohio State University, USA\\
\texttt{averly.1@buckeyemail.osu.edu,ning.104@osu.edu}\\
}

\maketitle

\begin{abstract}

Clinical named entity recognition (NER) aims to retrieve important entities within clinical narratives. Recent works have demonstrated that large language models (LLMs) can achieve strong performance in this task. While previous works focus on proprietary LLMs, we investigate how open NER LLMs, trained specifically for entity recognition, perform in clinical NER. Our initial experiment reveals significant contrast in performance for some clinical entities and how a simple exploitment on entity types can alleviate this issue. In this paper, we introduce a novel framework, entity decomposition with filtering, or EDF. Our key idea is to decompose the entity recognition task into several retrievals of entity sub-types and then filter them. Our experimental results demonstrate the efficacies of our framework and the improvements across all metrics, models, datasets, and entity types. Our analysis also reveals substantial improvement in recognizing previously missed entities using entity decomposition. We further provide a comprehensive evaluation of our framework and an in-depth error analysis to pave future works.

\end{abstract}

\section{Introduction}
\label{sec_intro}

Clinical narratives hold immense value for clinical experts~\cite{tayefi2021challenges, mahbub2022unstructured, raghavan2014essential, rannikmae2021developing}, largely due to their wealth of information often inaccessible in the structured data of the electronic health records (EHR)~\cite{mahbub2023question, goodman2022natural, kharrazi2018value, rannikmae2020accuracy, hernandez2019real, boag2018s}. Their free format, however, causes significant challenges for healthcare systems to utilize. The richness of information trapped within the narratives, followed by its significance, has spurred a plethora of works in tackling the \textit{clinical information extraction} problem within the clinical NLP community~\cite{wang2018clinical, landolsi2023information}. 

One key building block in clinical information extraction is named entity recognition (NER), focusing on identifying clinical concepts within these narratives. Prior methods~\cite{wang2018clinical} rely on either traditional natural language processing (NLP) techniques or supervised learning methods. Nevertheless, the former approach can be fragile, while the latter requires significant effort to annotate. In addition, supervised methods cannot simply scale for the large number of concepts available in the clinical domain~\cite{bodenreider2004unified}.

In light of this, Large Language Models (LLMs), with their strong capabilities for zero- and few-shot learning~\cite{chowdhery2023palm, brown2020language, thoppilan2022lamda, touvron2023llama}, serve as promising solutions for clinical NER. While previous works focus on LLMs trained on general tasks~\cite{agrawal2022large, liu2023deid, gero2023self}, here we focus on LLMs specifically trained for entity recognition, or \textit{open NER LLMs}~\cite{zhou2023universalner, ding2024rethinking}. Inspired by their results in clinical domain~\cite{zhou2023universalner}, outperforming even proprietary LLMs~\cite{brown2020language}, we conduct deeper investigations in this study. Surprisingly, our preliminary experiment (\autoref{subsec_prelim}) suggests a stark performance gap between retrieving different clinical entities (\autoref{fig_prelim}). For instance, an open NER LLM called \mbox{UniversalNER}~\cite{zhou2023universalner} performs significantly better at extracting medications rather than clinical treatments ($85.88\%$ vs $53.81\%$ Exact Match F1-scores). Upon closer inspection, we find these unidentified treatment entities can be effectively recognized by exploiting \textit{simpler, albeit specific}, entity types. For example, by explicitly specifying ``medication'' rather than ``treatment'' as input, the model can capture a substantial portion of the previously unidentified medication-related treatment entities. 

\begin{figure*}
\centering
\includegraphics[width=1\linewidth]{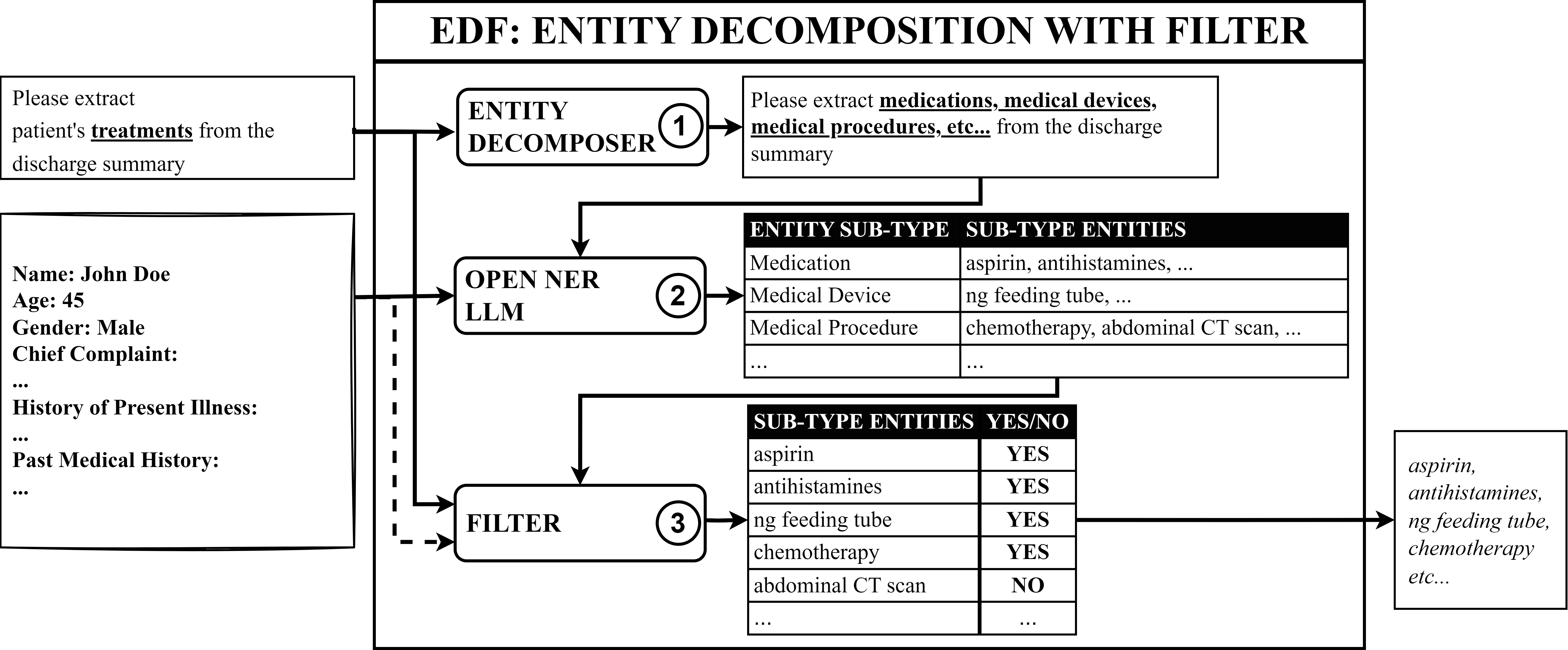}
\caption{\small \textbf{\OursLong}. Our novel framework is composed of three components: (1) \textbf{Entity Decomposer} breaks down the target entity type (e.g., treatment) into several entity sub-types (e.g., medication, medical device, medical procedure, etc), (2) \textbf{Open NER LLM} generates the sub-type entities, (3) \textbf{Filter} removes sub-type entities outside the target entity type. See~\autoref{sec_method} for details.}
\vskip -15pt
\label{fig_method}
\end{figure*}

Building upon this insight, we present a novel framework, \textit{entity decomposition with filtering}, or \Ours, aimed at tackling clinical NER. To the best of our knowledge, we are the first to explore strategies to effectively use open NER LLMs in the clinical domain \textit{without using any samples}. We draw inspiration from the divide-and-conquer paradigm~\cite{10.5555/280635}, which breaks down a complex problem into simpler sub-problems. Concretely, we posit that a direct retrieval of entities may be too complex for the model and instead propose to disentangle it into a series of retrievals through \textit{entity decomposition}. Unlike the previous approach~\cite{xie2023empirical}, entity decomposition breaks down the task by identifying through their \textit{entity sub-types}, which, ideally, are easier to retrieve. Nonetheless, entity decomposition alone is insufficient since some entity sub-types do not form strict subsets (further discussed in~\autoref{filter}). To address this, we introduce a \textit{filtering} mechanism in our framework to further improve performance. We illustrate them in~\autoref{fig_method}.

Our work introduces a cost-effective approach to improve clinical entity recognition using open-source large language models. Overall, we observe improvements across models, metrics, datasets and entity types. Our ablation study also reveals the robustness of our framework, allowing users to adjust the components based on their performance and cost. 

\section{Related Work}

\subsection{LLMs for Clinical NER}
\label{sec_related_1}

LLMs are promising for many clinical tasks~\cite{singhal2022large, agrawal2022large, clusmann2023future}. Concurrently, several works aim to improve their performance on clinical NER. For instance, Agrawal \emph{et.al.}~\cite{agrawal2022large} proposes a guided prompt design along with a resolver to handle the structured output space required by NER, while others~\cite{hu2024zero, hu2023zero, liu2023deid} use prompt engineering. Outside the clinical domain, several works tackle NER either by framing it as a sequence labeling task~\cite{wang2023gpt}, using label decomposition and syntactic augmentation~\cite{xie2023empirical}, or improving the structured label space~\cite{li2024simple}, similar to Agrawal \emph{et.al.}~\cite{agrawal2022large}. Most of these works focus on LLMs trained in handling diverse tasks such as ChatGPT~\cite{brown2020language}. In contrast, we focus on open NER LLMs~\cite{zhou2023universalner, ding2024rethinking}, which have two key differences. First, they are trained specifically for entity recognition tasks and do not require structured output space handling~\cite{agrawal2022large, li2024simple}. Second, their instruction-tuning mostly focused on the diversity of entities rather than the instructions (e.g., keeping the prompt constant), which may limit the efficacy of prompt engineering techniques. Furthermore, unlike previous works~\cite{hu2024zero, hu2023zero, liu2023deid}, \textit{our work does not fall under prompt engineering}. Notably, prompt engineering is limited to prompt-based models, while our work is model-agnostic and, thus, is applicable to BERT-based models~\cite{zaratiana2023gliner}.


\subsection{Task Decomposition in LLMs}

The idea of task decomposition, solving complex tasks through solving its constituent simpler sub-tasks, can be dated back to~\cite{lazarou1998incidence}. Previous works propose task decompositions for LLMs to tackle complex problems~\cite{zhou2022least, xie2023empirical}. Concurrent with our work, Xie \emph{et.al.}~\cite{xie2023empirical} suggests decomposing NER into a multi-turn dialogue, asking the model one question for each label. However, some open NER LLMs~\cite{zhou2023universalner} can only extract one label at a time, thus limiting the efficacy of Xie \emph{et.al.}~\cite{xie2023empirical}. Here, \textit{we propose to decompose NER on entity-level rather than label-level}. Concretely, we can further decompose each label into simpler labels. Our method also complements Xie \emph{et.al.}~\cite{xie2023empirical} since these decompositions can be performed sequentially. Besides, our work aims to improve open NER LLMs, which have several key differences from other LLMs as briefly discussed in~\autoref{sec_related_1}

\subsection{Open NER LLMs}

Clinical narratives fall under domains with a large number of concepts and scarce annotations. Thus, developing open named entity recognition LLMs~\cite{zhou2023universalner, ding2024rethinking} is timely and crucial research for clinical NER. Despite the progress, existing works focus on training the backbone models. Furthermore, these models present a unique challenge and cannot be treated similarly to other LLMs (\autoref{sec_related_1}). Our work paves a way to adapt them for clinical domains without finetuning.


\section{Method}
\label{sec_method}

\subsection{Problem Definition}
\label{subsec_problem}

Clinical narrative holds important entities about a patient's medical history. In this work, we aim to tackle clinical NER, focusing on extracting them. We frame the problem through the lens of a text-generative task. Let $\bx$ be a clinical narrative, and further let $\bt$ be the \textit{target entity type} we want to extract. We aim to retrieve the \textit{target entities set} $\sY$ corresponding to $\bt$ from $\bx$. To illustrate, if $\bx$ is a patient's discharge summary and $\bt=\textnormal{``medication''}$, then the goal is to extract medication entities in the discharge summary. In this case, the ouput can be $\sY=\{\textnormal{``aspirin'', ``methanol'', ...}\}$.

\subsection{\OursLong}
\label{subsec_edf}

As introduced in~\autoref{sec_intro}, directly retrieving the target clinical entities may be too challenging, particularly for models without domain-specific training, such as open NER LLMs. We propose to break the task into multiple retrievals of \textit{sub-type entities} instead. We define sub-type entities as entities belonging to a a sub-type of the target entity type. Concretely, let $\shY_i$ be a set of sub-type entities corresponding to the $i$-th \textit{entity sub-type} $\bs_i$ and let $\shY$ be our predicted complete set, where ideally $\shY_i \subseteq \shY$. The first part of our framework, \textit{entity decomposer}, aims to iteratively collect $\shY_i$ to produce $\shY$, $\smash{\bigcup\nolimits^{N}_{i=1} \shY_i = \shY}$. The last part of our framework, \textit{filter}, involves removing sub-type entities in $\shY$ outside the target entity type $t$. \mbox{The filtered version $\shY_f$} then serves as the final output. That is, $\sY = \shY_f$ We provide more details of our framework in the following sections.~\autoref{fig_method} presents the overall architecture of our framework, entity decomposition with filtering, or EDF.

\subsubsection{Entity Decomposer}
\label{decomposer}

The first step in our framework is to identify what constitutes entity sub-types for the target entity type $t$. Let $\sD(\bt) = \sS$ be the \mbox{\textit{entity decomposer}} module, aimed to produce \textit{a set of} $N$ \textit{entity sub-types} $\sS = \{\bs_1, \bs_2, ..., \bs_N\}$ using the target entity type $\bt$. To illustrate how the module works, let's take ``treatment'' as an example. In a patient's discharge summary, ``treatment'' constitutes a myriad of entities, including medications, medical procedures, and more (see~\autoref{fig_method}). In this case, $\sD$ aims to decompose $\bt = \textnormal{``treatment''}$ into $\sS=\{\textnormal{``medication'', ``medical procedure'', ...}\}$.

There are several ways to construct this module, and clinical practitioners can resort to different methods based on costs and performances. Some examples include manually curating the entity sub-types (possibly involving clinical experts) or obtained using existing tools such as a medical knowledge base~\cite{bodenreider2004unified}. We provide details in~\autoref{app_entity_decomposer}

\subsubsection{Open NER LLM}
\label{openner_llm}

After defining the entity sub-types $\sS$, the next step is to retrieve the sub-type entities $\shY$. In this work, we leverage an \textit{open NER LLM}. We base its formal definition on previous works~\cite{zhou2023universalner}. Let $R$ be an open NER LLM tasked to retrieve the sub-type entities $\shY_i$ of $\bs_i$ from the clinical narrative $\bx$. That is, \mbox{$R(\bx, \bs_i)=\shY_i$}. To construct the complete set $\shY$, the model would iteratively collect $\shY_i$ from each $\bs_i$. This may be of concern, given that it requires multiple iterations. To address this, we also consider other variants~\cite{ding2024rethinking, zaratiana2023gliner} of open NER LLMs, which we formally define as $R^*(\bx, \sS)=\shY$. In contrast to $R$, these variants are capable of simultaneously extracting multiple entity types at once, thereby reducing the whole iterative sub-type entity retrievals to only one forward pass.

We base the use of an open NER LLM on several key reasons. First, open NER LLMs are versatile in recognizing arbitrary entities, which is a characteristic of sub-type entities (e.g. they are not bounded to a predefined label set). Second, in contrast to other LLMs, open NER LLMs are explicitly trained for NER tasks, substantially reducing the effort of mapping LLM generative output to the structured output space of NER~\cite{agrawal2022large}. In addition, our preliminary experiment in~\autoref{subsec_prelim} suggests that open NER LLMs perform surprisingly well in recognizing basic clinical entities such as medications, making them strong candidates for this module.

\subsubsection{Filter}
\label{filter}

We discuss the last step of our framework here. Formally, let $f(\shY,\bt, C) = \shY_f$ be the \textit{filter} module, where $C$ denotes a \textit{context} and $\shY_f \subseteq \shY$. In this framework, $f$ can be viewed as a binary classifier, assigning a positive label if the entity corresponds to $t$, and a negative label otherwise. The filter module then aggregates and outputs the positively classified entities. Overall, $f$ aims to eliminate sub-type entities within $\shY$ that do not fall under the target entity type $t$ based on its context $C$ (e.g. the paragraph the entity exists). Take ``treatment'' and ``medical procedure'' for example. While ``treatment'' includes ``medical procedure'', not all ``medical procedure'' qualify as a ``treatment''. For instance, some medical procedures (e.g., endoscopy) may serve purely for diagnosis purposes and should not be classified as ``treatment.'' 

We provide the rationale behind using a context $C$ in the filter module. Unlike general domain entities, clinical entities can largely rely on context cues. To illustrate, consider ``adverse drug event (ADE)''~\cite{zed2008incidence, lazarou1998incidence}. By definition, ``ADE'' is an \textit{"injury resulting from a medical intervention"}~\cite{henry20202018}. Thus, one of its sub-types may be ``injury''. However, for a filter to dictate whether an injury corresponds to ``ADE'', it needs to be aware of where the injury comes from. Injuries due to an accidental fall may not be an ADE if the patient does not have any medical interventions. In other words, \textit{the filter requires the context in which the entity occurs to provide an accurate prediction}. Moreover, our experiment suggests that while some clinical entities do not need contextual information, they can benefit from it (see~\autoref{subsubsec_result_filter_context}). 

\section{Experimental Setup}
\label{sec_exp_setup}

We provide the experimental setup here and leave the details in the Appendix. All of the base models are available in huggingface\footnote{\url{https://huggingface.co}}.

\subsection{Open NER LLMs}
\label{subsec_exp_openner_llm}

We take SOTA open NER LLMs in our experiments and further improve them. Per definition in~\autoref{openner_llm}, they may be categorized based on how many entity types can be extracted simultaneously. To this end, we use \textbf{\mbox{UniversalNER}}~\cite{zhou2023universalner} and \textbf{GNER}~\cite{ding2024rethinking} as the representative for $R$ and $R^*$, respectively. Concretely, we use \mbox{\textsc{UniversalNER-type-7B}} and \mbox{\textsc{GNER-LLaMA-7B}}. Both are finetuned on the PileNER dataset~\cite{zhou2023universalner}, which is generated from GPT 3.5~\cite{brown2020language}. We only experiment with their default prompts since, in contrast to other LLMs, open NER LLMs are trained on a set of diverse entity types rather than prompts~\cite{zhou2023universalner}

\subsection{Entity Decomposers}
\label{subsec_exp_decomposer}

Here, we experiment with different techniques to decompose entities. \textbf{First}, given that clinical narratives require specialized knowledge, we consider entity sub-types curated by clinical experts. Specifically, we take the annotation guidelines available from the datasets. \textbf{Second}, we use \mbox{ChatGPT}~\cite{brown2020language} to decompose clinical entities automatically. We draw our inspiration from the recent success of ChatGPT in the clinical domain~\cite{agrawal2022large, singhal2022large}. Furthermore, using ChatGPT for entity decomposition is more cost-effective and scalable. \textbf{Third}, we utilize the Unified Medical Language System (UMLS)~\cite{bodenreider2004unified}, a medical knowledge base, for retrieving entity sub-types. We provide more details in the~\autoref{app_entity_decomposer}.

\subsection{Filters}
\label{subsec_exp_filter}

We use \textbf{Asclepius}~\cite{kweon2023publicly} and \textbf{LlaMA-2}~\cite{touvron2023llama} trained on clinical and general domains, respectively. Specifically, we use \mbox{\textsc{Asclepius-Llama2-7B}} and \mbox{\textsc{LlaMA-2-Chat-7B}} versions. Given the inherent generative nature of LLMs, we restrict their outputs to "Yes/No" responses (when applicable) using grammar-constrained decoding~\cite{geng2023flexible}. This strategic constraint reduces the number of generated tokens, resulting in increased inference speed. By default, we prompt the model by asking ``\textit{Can \{entity\} be considered as a/an \{entity\_type\}?}''. We try different prompts in~\autoref{subsubsec_result_filter_prompt}. For entities that require context, we use a simple preprocessing method so that the context provides sufficient information to extract the clinical entities. That is, we include the whole paragraph (or more) in which the entity occurs from the clinical narrative.

\subsection{Datasets}
\label{subsec_exp_dataset}

We focus our experiments on extracting concepts from publicly available clinical notes\footnote{All datasets are available from the providers under appropriate data usage agreements}. We use \textbf{\mbox{ClinicalIE}}~\cite{agrawal2022large}, \textbf{\mbox{i2b2 2010}}~\cite{uzuner20112010}, \textbf{i2b2 2012}~\cite{sun2013evaluating}, \textbf{i2b2 2018 Task 2}~\cite{henry20202018} and \textbf{CLEF 2014}~\cite{mowery2014task} datasets in this paper. They are available in Harvard DBMI\footnote{\url{https://portal.dbmi.hms.harvard.edu/}} for i2b2 datasets, PhysioNet\footnote{\url{https://physionet.org/}} \cite{goldberger2000physiobank} for CLEF 2014 and huggingface for ClinicalIE. In these datasets, context is not required to identify clinical entities except for the i2b2 2018 dataset. We provide further details in the~\autoref{app_datasets}.

\subsection{Baselines and Metrics}
\label{subsec_exp_baseline}

Given the scarce methods to compare with, we use the method developed in Xie \emph{et.al.}~\cite{xie2023empirical} as our baseline. Concretely, we extract each entity type one at a time using UniversalNER and GNER. We also compare with \textsc{UniversalNER-all}, trained with both PileNER and over 40 supervised datasets, including our experiment sets. We use \textit{Precision} (P), \textit{Recall} (R), and \textit{Exact Match F1-Score} (F1) as evaluation metrics, similar to previous works.

\section{Experimental Results}
\label{sec_exp_result}

Throughout this section, we abbreviate the entity types in the tables as follows to save space: \texttt{Tr} for treatment, \texttt{Pr} for problem, \texttt{Te} for test, \texttt{CD} for clinical department, \texttt{DD} for disease/disorder, \texttt{AD} for adverse drug, and \texttt{ADE} for adverse drug event.

Given the space limit, we include more results in the Appendix, including few-shot (\autoref{app_fewshot}), performance on a BERT-based model (\autoref{app_non_llm}), and error analysis on \texttt{CD} performance drop (\autoref{app_drop_cd}). 

\subsection{Preliminary Experiment}
\label{subsec_prelim}

We conduct a preliminary experiment to confirm that open NER LLMs perform better at recognizing sub-type entities rather than the target entity types. For target entities, we use the i2b2 2012 dataset, which contains decomposable entity types (i.e., entities can further be divided into sub-type entities). For sub-type entities, we use ClinicalIE, a medication extraction dataset. 

\begin{figure}[!t]
\centering
\includegraphics[width=0.8\linewidth]{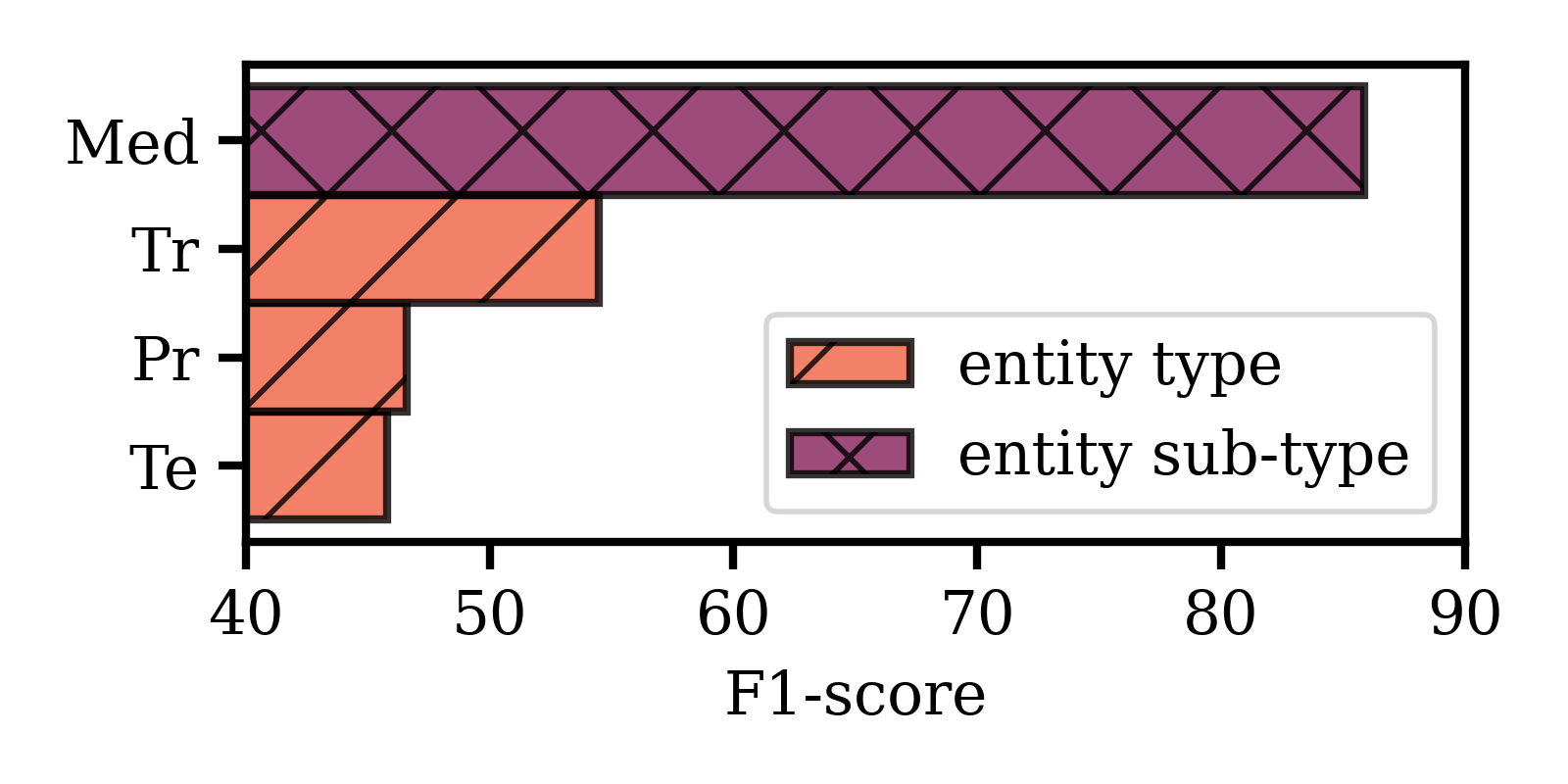}
\vskip -5pt
\caption{\small Open NER LLM (UniversalNER) performs better at extracting entity sub-type (\texttt{Med}) rather than the target entity type (\texttt{Tr}, \texttt{Pr}, \texttt{Te}). We use ClinicalIE for \texttt{Med} (medication) and i2b2 2012 for \texttt{Tr} (treatment), \texttt{Pr} (problem), and \texttt{Te} (test). Discussion in~\autoref{subsec_prelim}}
\vskip -15pt
\label{fig_prelim}
\end{figure}
~\autoref{fig_prelim} illustrates the result and confirms our hypothesis. That is, it is harder to recognize the target entity types (that are decomposable) compared to the sub-type entities. For instance, we observe a stark difference between ``medication'' extraction and ``treatment'' extraction, where the former is an entity sub-type of treatment.

\begin{table*}[!t]
  \caption{F1-score performance ($\%$) comparison between baseline (B) from Xie \emph{et.al.}~\cite{xie2023empirical}, Entity Decomposer (ED) only, Filter (F) only, supervised (UniNER-all) and \Ours~(Ours) methods across datasets, entity types, and models. We use Asclepius and the default prompt strategy (\autoref{filter}) for the filter and annotation guideline for the entity decomposer. We mark the best results in \textbf{bold} and second-best in \underline{underlined}. Discussion in~\autoref{subsec_overall_perf}}
  \label{table_main}
  \centering
  \footnotesize
  \begin{threeparttable}
      \begin{tabular}{
        @{\hspace{2pt}}l@{\hspace{4pt}}
        @{\hspace{4pt}}l@{\hspace{4pt}}
        @{\hspace{4pt}}r@{\hspace{5pt}}
        @{\hspace{5pt}}r@{\hspace{5pt}}
        @{\hspace{5pt}}r@{\hspace{5pt}}
        @{\hspace{5pt}}r@{\hspace{5pt}}
        @{\hspace{4pt}}r@{\hspace{5pt}}
        @{\hspace{5pt}}r@{\hspace{5pt}}
        @{\hspace{5pt}}r@{\hspace{5pt}}
        @{\hspace{5pt}}r@{\hspace{5pt}}
        @{\hspace{5pt}}r@{\hspace{5pt}}
        @{\hspace{4pt}}r@{\hspace{3pt}}
        @{\hspace{6pt}}c@{\hspace{2pt}}
      }
      \toprule

      \multirow{2}{*}{Dataset} & \multirow{2}{*}{\makecell[l]{Entity\\Type}} & \multicolumn{5}{c}{UniNER} & \multicolumn{5}{c}{GNER} & \multirow{2}{*}{\makecell{UniNER-all\\(Supervised)}} \\
      \cmidrule(r){3-7}\cmidrule(r){8-12}
      && B & ED & F & EDF & $\Delta$ & B & ED & F & EDF & $\Delta$ & \\
      \midrule
      \multirow{3}{*}{i2b2 2010} & \texttt{Tr} & 53.81 & 43.56 & \underline{54.35} & \textbf{59.63} & $+$5.82 & 53.31 & 39.43 & \underline{62.72} & \textbf{63.23} & $+$9.92 & \textit{74.95} \\
      & \texttt{Pr} & 49.71 & 41.99 & \textbf{52.77} & \underline{51.46} & $+$1.75 & 40.40 & 35.91 & \textbf{50.77} & \underline{50.62} & $+$10.22 & \textit{73.11} \\
      & \texttt{Te} & \textbf{48.78} & 36.99 & 40.58 & \underline{44.17} & $-4.61$ & \underline{37.27} & 36.40 & 37.23 & \textbf{45.78} & $+$8.51 & \textit{72.43} \\
      \midrule
      \multirow{4}{*}{i2b2 2012} & \texttt{Tr} & \underline{54.49} & 48.25 & 52.82 & \textbf{58.25} & $+$3.76 & 50.38 & 40.16 & \underline{55.63} & \textbf{57.09} & $+$6.71 & \textit{72.37} \\
      & \texttt{Pr} & 46.61 & 42.45 & \underline{49.64} & \textbf{50.24} & $+$3.63 & 41.11 &  38.00 & \underline{47.15} & \textbf{48.82} & $+$7.71 & \textit{75.16} \\
      & \texttt{Te} & \underline{45.78} & 31.52 & 43.98 & \textbf{46.34} & $+$0.56 & 33.17 & 27.55 & \underline{41.09} & \textbf{46.66} & $+$13.49 & \textit{65.47} \\
      & \texttt{CD} & \textbf{41.44} & 32.57 & 34.51 & \underline{38.66} & $-$2.78 & \textbf{58.88} & 20.46 & \underline{39.20} & 37.74 & $-$21.14 & \textit{44.37} \\
      \midrule
      CLEF 2014 & \texttt{DD} & \underline{46.73} & 45.94 & 45.26 & \textbf{58.25} & $+$11.52 & 18.06 & \underline{21.11} & 18.44 & \textbf{26.91} & $+$8.85 & \textit{63.19} \\
      \midrule
      \multirow{2}{*}{i2b2 2018} & \texttt{AD}$^{\dagger\ddagger}$ & 19.07 & 8.59 & \textbf{27.36} & \underline{24.92} & $+$5.85 & 3.16 & 4.54 & 10.08 & 13.66 & $+$10.50 & \textit{14.37} \\
      & \texttt{ADE}$^{\dagger}$ & \underline{9.56} & 2.64 & \textbf{15.90} & 9.43 & $-$0.13 & 0.61 & 1.09 & \underline{2.40} & \textbf{4.05} & $+$3.44 & \textit{31.18} \\
      \midrule
      Avg. && 41.60 & 33.45 & \underline{41.72} & \textbf{44.14} & $+$2.54 & 33.64 & 26.47 & \underline{36.47} & \textbf{39.46} & $+$5.82 & \textit{58.66} \\
      \bottomrule
      \end{tabular}

      \begin{tablenotes}[normal,flushleft]
          \begin{footnotesize}
          \item[]
      $\dagger$ \texttt{AD} and \texttt{ADE} are entity types that require context $\ddagger$ UniNER-all is not trained to extract \texttt{AD} entities
          \par
          \end{footnotesize}
      \end{tablenotes}
  \end{threeparttable}
  \vskip -15pt
\end{table*}

\subsection{Overall Performance}
\label{subsec_overall_perf}
 
\begin{figure}[t]
\centering
\includegraphics[width=1\linewidth]{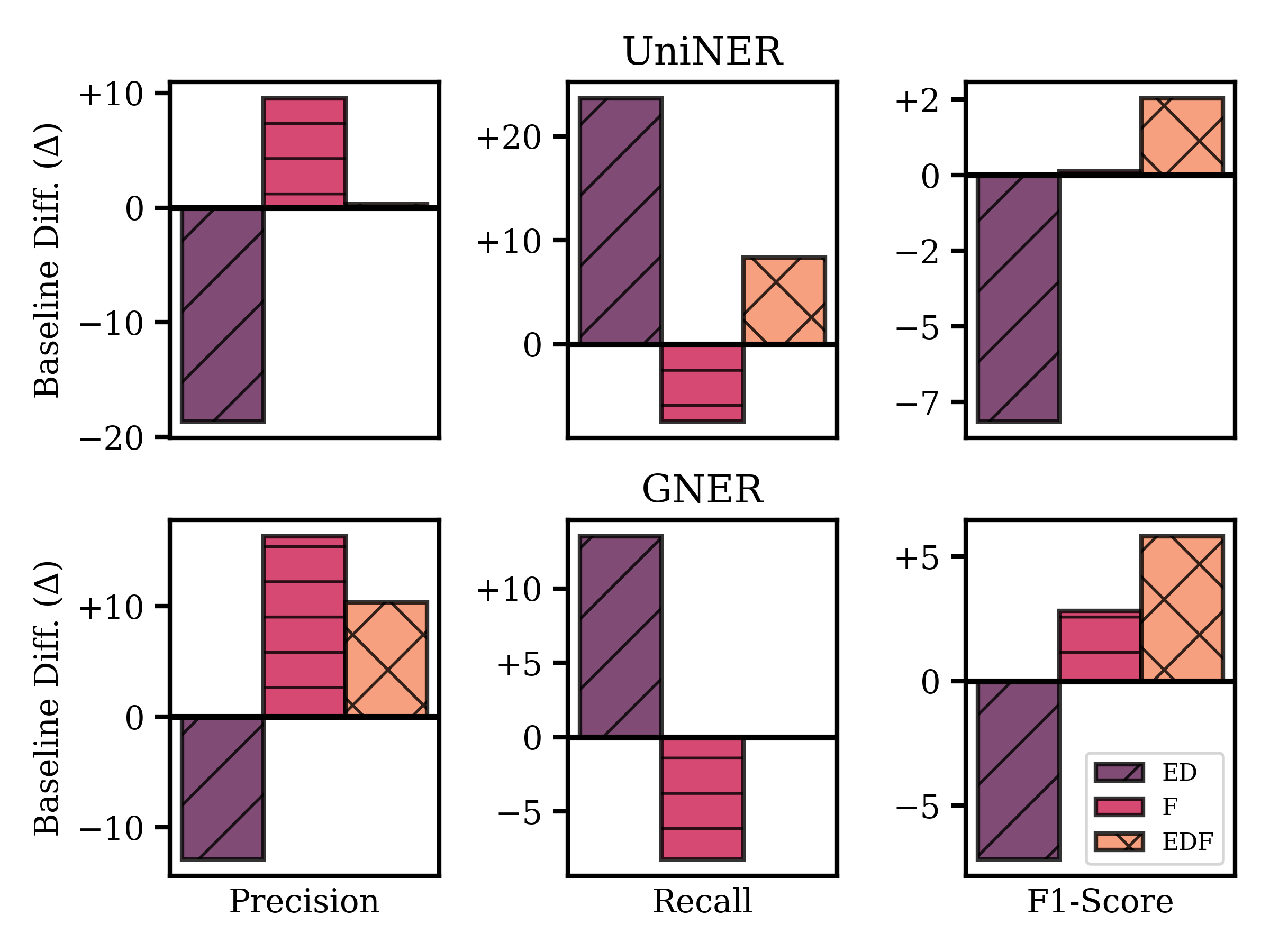}
\vskip -5pt
\caption{\small Average performance improvement to baseline across metrics and models. Entity decomposition (ED) improves recall but degrades precision. Filter (F) increases precision but decreases recall. Our method (EDF) achieves better performance overall. Discussion in~\autoref{subsec_overall_perf}}
\vskip -15pt
\label{fig_prec_rec_f1}
\end{figure}

We present our results in~\autoref{table_main} and~\autoref{fig_prec_rec_f1}. For detailed numbers on precision and recalls, we leave them in the~\autoref{app_recall_precision}.

\textbf{\textit{On average, EDF outperforms baseline by 2.54\% and 5.82\% F1-score on UniversalNER and GNER, respectively.}} Interestingly, for some entity types (e.g. treatments and tests in i2b2 2010), GNER performs similarly or even outperforms \mbox{UniversalNER}. This suggests that models that can recognize multiple entities simultaneously can benefit more from using our framework.

\textbf{\textit{Entity decomposition (ED) improves recall but decreases precision}}. As illustrated in~\autoref{fig_prec_rec_f1}, we observe a consistent improvement in recall across diverse datasets and entity types for both models, suggesting that entity decomposition facilitates the identification of previously missing entities while being robust to the backbone models. For precision, however, we notice a drop in performance. As discussed in~\autoref{filter}, some sub-type entities may not form a subset of the target entities, causing performance degradation on precision. Further examination of the F1-score reveals a decline in overall performance, which justifies the necessity of incorporating a filtering mechanism.

\textbf{\textit{Conversely, filtering (F) benefits precision but degrades recall}}. This shows contrasting results compared to entity decomposition across datasets, entity types, and models. The overall F1-score improvement on EDF compared to using each component individually suggests that entity decomposition and filtering complement each other. This emphasizes the necessity to incorporate both of them.

\textbf{\textit{EDF is robust to out-of-distribution entities compared to supervised training}}. We want to emphasize that \textit{our method does not require any training} as opposed to the supervised approach. We use UniversalNER+EDF and UniversalNER-all as comparison. Despite the performance gap, we observe that on entities not covered in the training label set (e.g., adverse drug or \texttt{AD}), EDF outperforms by more than $10\%$ on the F1-score. This shows the robustness of our method.

\begin{figure*}[!t]
\centering
\includegraphics[width=1\linewidth]{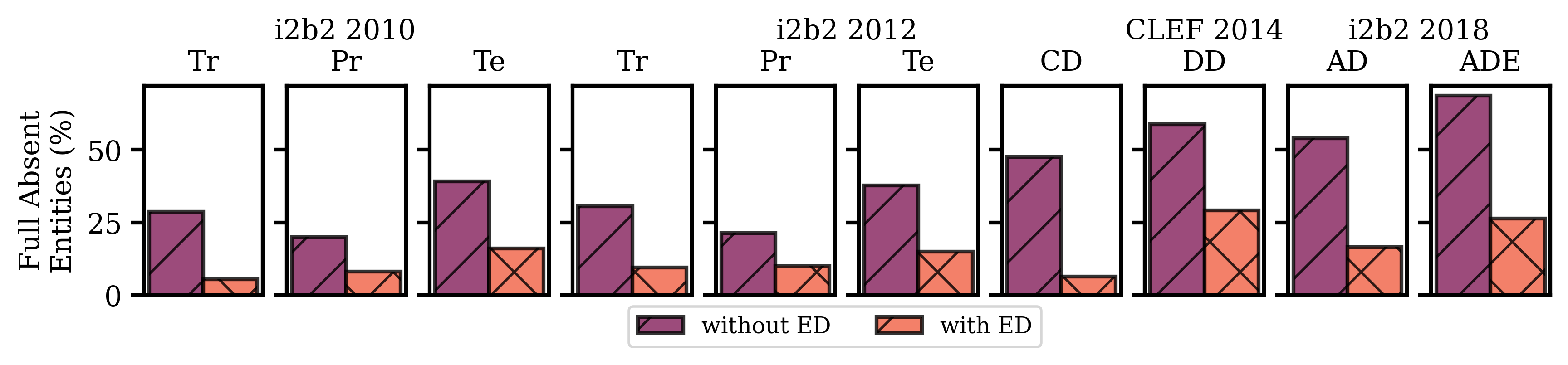}
\vskip -5pt
\caption{\small Entity Decomposition (ED) captures most previously fully absent entities. Lower value is better. Discussion in~\autoref{subsubsec_error1}}
\vskip -15pt
\label{fig_fully_absent}
\end{figure*}

\subsection{Ablation Study}

We perform additional experiments to test the efficacy of our framework using different entity decomposers or filters. We focus on the overall performance or F1-score. To reduce the cost of our experiments, we only experiment with the i2b2 2012 dataset, given their diversity in entity types. Unless otherwise specified, we use UniNER as the open NER LLM, annotation guideline for the entity decomposer module, and Asclepius for the filter.

\subsubsection{Entity Decomposers}

First, we conduct ablation study with different entity decomposers as described in~\autoref{subsec_exp_decomposer} and present our results in~\autoref{table_decomposer}. We do not experiment with ChatGPT and UMLS for the ``clinical department'' entity type since (1) ChatGPT is unable to produce reasonable entity sub-types and (2) we find there are no correspondence semantic types in UMLS. 

\textbf{\textit{EDF is robust to entity decomposer module}}. We observe competitive performance between a general LLM, medical knowledge base and clinical experts. For instance, on treatment entities, the performance between clinical experts and ChatGPT is $58.25\%$ vs $58.09\%$. Thus, even without curation from clinical experts, our method can achieve competitive results.

\textbf{\textit{Without filters, UMLS outperforms other entity decomposers}}. Most entity sub-types in UMLS form subsets to the target entities; hence, a filter may not be necessary in contrast to other entity decomposers. Interestingly, for some entities, it performs better than vanilla prompt (e.g. treatments and problems). This is significant since even without filtering, our framework, specifically entity decomposition, can still outperform the vanilla approach with proper curation of entity sub-types. We remark that this does not take away the value of a filter module. For instance, UMLS can benefit more from integrating it, as shown in~\autoref{table_decomposer}.

\subsubsection{Filter Models}

\begin{table}[!t]
  \caption{Ablation Study on Entity Decomposer.}
  \label{table_decomposer}
  \centering
  \footnotesize
  \begin{threeparttable}
      \begin{tabular}{
        @{\hspace{6pt}}l@{\hspace{4pt}}
        @{\hspace{4pt}}c@{\hspace{4pt}}
        @{\hspace{4pt}}c@{\hspace{6pt}}
        @{\hspace{6pt}}c@{\hspace{6pt}}
        @{\hspace{6pt}}c@{\hspace{6pt}}
        @{\hspace{6pt}}c@{\hspace{6pt}}
      }
      \toprule

      \multirow{2}{*}{\makecell[l]{Entity\\Decomposer}} & \multirow{2}{*}{Filter} & \multicolumn{4}{c}{Entity Type} \\
      \cmidrule{3-6}
      && \texttt{Tr} & \texttt{Pr} & \texttt{Te} & \texttt{CD} \\
      \midrule
      Baseline && 54.49 & 46.61 & 45.78 & \textbf{41.44} \\
      \midrule
      Annotation & \multirow{3}{*}{\xmark} & 48.25 & 42.45 & 31.52 & 32.57 \\
      ChatGPT && 48.62 & 44.05 & 43.80 & - \\
      UMLS && 55.57 & 46.98 & 44.68 & - \\
      \midrule
      Annotation & \multirow{3}{*}{\cmark} & \textbf{58.25} & 50.24 & \textbf{46.34} & 38.66 \\
      ChatGPT && 58.09 & \textbf{51.12} & 44.83 & - \\
      UMLS && 58.23 & 51.04 & 44.71 & - \\
      
      \bottomrule
      \end{tabular}
  \end{threeparttable}
  \vskip -5pt
\end{table}

\begin{table}[!t]
  \caption{Ablation Study on Filter Model.}
  \label{table_filter_model}
  \centering
  \footnotesize
  \begin{threeparttable}
      \begin{tabular}{
        @{\hspace{3pt}}c@{\hspace{2pt}}
        @{\hspace{2pt}}l@{\hspace{4pt}}
        @{\hspace{4pt}}c@{\hspace{5pt}}
        @{\hspace{5pt}}c@{\hspace{5pt}}
        @{\hspace{5pt}}c@{\hspace{5pt}}
        @{\hspace{5pt}}c@{\hspace{3pt}}
      }
      \toprule

      \multirow{2}{*}{\makecell[l]{Entity\\Decomposer}} & \multirow{2}{*}{\makecell[l]{Filter\\Model}} & \multicolumn{4}{c}{Entity Type} \\
      \cmidrule{3-6}
      && \texttt{Tr} & \texttt{Pr} & \texttt{Te} & \texttt{CD} \\
      \midrule
      \makecell[l]{Baseline} && 54.49 & 46.61 & 45.78 & \textbf{41.44} \\
      \midrule
      \multirow{2}{*}{\xmark} & Asclepius & 52.82 & 49.64 & 43.98 & 34.51 \\
      & Llama2 & 51.90 & 47.49 & 45.27 & 35.37 \\
      \midrule
      \multirow{2}{*}{\cmark} & Asclepius & \textbf{58.25} & \textbf{50.24} & \textbf{46.34} & 38.66 \\
      & Llama2 & 53.94 & 45.55 & 37.41 & 38.69 \\     
      
      \bottomrule
      \end{tabular}
  \end{threeparttable}
  \vskip -15pt
\end{table}

Here, we investigate how different filter models affect the overall performance of our framework. Specifically, we compare domain-specific and general LLMs. We present the results in~\autoref{table_filter_model}

\textbf{\textit{Clinical model is better at recognizing entities requiring clinical expertise}}. Specifically, we observe that it is superior to a general domain model in ``treatment'', ``problem'', and ``test'' entities. For ``clinical department'', however, they perform similarly. This is unsurprising since the former often requires clinical-specific knowledge compared to the latter.

\subsubsection{Filter Context}
\label{subsubsec_result_filter_context}

\begin{table}[!t]
  \caption{Ablation Study on Filter Context.}
  \label{table_filter_context}
  \centering
  \footnotesize
  \begin{threeparttable}
      \begin{tabular}{
        @{\hspace{3pt}}c@{\hspace{2pt}}
        @{\hspace{2pt}}l@{\hspace{4pt}}
        @{\hspace{4pt}}c@{\hspace{5pt}}
        @{\hspace{5pt}}c@{\hspace{5pt}}
        @{\hspace{5pt}}c@{\hspace{5pt}}
        @{\hspace{5pt}}c@{\hspace{3pt}}
      }
      \toprule

      \multirow{2}{*}{\makecell[l]{Entity\\Decomposer}} & \multirow{2}{*}{\makecell[l]{Filter\\Context}} & \multicolumn{4}{c}{Entity Type} \\
      \cmidrule{3-6}
      && \texttt{Tr} & \texttt{Pr} & \texttt{Te} & \texttt{CD} \\
      \midrule
      \makecell[l]{Baseline} && 54.49 & 46.61 & 45.78 & \textbf{41.44} \\
      \midrule
      \multirow{3}{*}{\xmark} & none & 52.82 & 49.64 & 43.98 & 34.51 \\
      & sentence & 53.85 & 42.90 & 43.91 & 37.18 \\
      & document & 54.67 & 41.53 & 44.45 & 44.91 \\
      \midrule
      \multirow{3}{*}{\cmark} & none & \textbf{58.25} & \textbf{50.24} & 46.34 & 38.66 \\
      & sentence & 60.58 & 45.17 & \textbf{46.68} & 40.80 \\
      & document & 58.24 & 42.24 & 45.31 & \textbf{48.90} \\  
      \bottomrule
      \end{tabular}
  \end{threeparttable}
  \vskip -10pt
\end{table}

As discussed in~\autoref{filter}, some clinical entities need context. Here, we investigate whether context helps for entities not requiring it. We compare the performance between filtering (1) without context, (2) including the sentence the entity appears in, and (3) providing the whole clinical narrative or document. We show the results in~\autoref{table_filter_context}.

\textbf{\textit{Context may improve or hurt performance}}. Overall, we observe mixed results across different entity types, with and without an entity decomposer. For instance, we observe slight improvements for ``treatment'' and ``test'' entities. On the other hand, the performance of the ``problem'' entity consistently drops the more context we provide. We provide further analysis in~\autoref{subsubsec_error4}.

\subsubsection{Filter Prompts}
\label{subsubsec_result_filter_prompt}

\begin{table}[!t]
  \caption{Ablation Study on Filter Prompt}
  \label{table_filter_ent_desc}
  \centering
  \footnotesize
  \begin{threeparttable}
      \begin{tabular}{
        @{\hspace{4pt}}c@{\hspace{8pt}}
        @{\hspace{0pt}}c@{\hspace{8pt}}
        @{\hspace{8pt}}c@{\hspace{8pt}}
        @{\hspace{8pt}}c@{\hspace{8pt}}
        @{\hspace{8pt}}c@{\hspace{4pt}}
      }
      \toprule

      \multirow{2}{*}{Entity Description} & \multicolumn{4}{c}{Entity Type} \\
      \cmidrule{2-5}
      & \texttt{Tr} & \texttt{Pr} & \texttt{Te} & \texttt{CD} \\
      \midrule
      \xmark & \textbf{58.25} & \textbf{50.24} & \textbf{46.34} & 38.66 \\
      \cmark & 50.32 & 47.38 & 33.18 & \textbf{45.97} \\
      
      \bottomrule
      \end{tabular}
  \end{threeparttable}
  \vskip -15pt
\end{table}


LLMs are often brittle to prompting strategies~\cite{zhu2023promptbench}. We experiment with how incorporating the entity description into the filter affects our framework's performance. For ``treatment'', ``problem'' and ``test'', we use descriptions available in i2b2 2010 annotation guidelines. For ``clinical department'', we provide our own definition. We put the prompts in~\autoref{app_filter_prompt} and present the results in~\autoref{table_filter_ent_desc}.

\textbf{\textit{Complex entity description degrades performance}}. Our experiment shows a notable performance drop in all entity types except ``clinical department''. We hypothesize that the descriptions provided in the guidelines may be too complex for the model to understand. In contrast, we observe more than $7\%$ F1-score improvement on ``clinical department'', which uses our handcrafted and concise definition. 

\subsection{Error Analysis}
\label{subsec_error}

\subsubsection{Entity Decomposition Missing Entities}
\label{subsubsec_error1}

Despite the significant improvement in recall through entity decomposition, some entities remain unrecognizable. Thus, we analyze the potential sources of these errors. First, we calculate the percentage of entities \textit{fully absent} from the predictions. To illustrate, if the label is ``his aspirin'' and the prediction is ``aspirin'', we do not deem it fully absent since the prediction partially captures the label.~\autoref{fig_fully_absent} illustrates the percentage of fully absent entities for each entity type in the dataset. 

\textbf{\textit{Entity decomposition significantly reduces the number of fully absent entities}}. For instance, only $5.5\%$ entities are fully absent for ``treatment'' in the i2b2 2010 dataset after entity decomposition. We observe improvements across all entity types.

\textbf{\textit{The majority of fully absent entities are abbreviations and homonyms}}. For example, open NER LLM cannot capture ``CVA'', an abbreviation for ``cerebral vascular accident", after entity decomposition. Another example is ``delivery", which carries nuanced meanings in different contexts (e.g. childbirth or route of medications). Furthermore, certain entities such as ``HD'' are both abbreviations and homonyms (i.e., high definition vs hemodialysis).

\subsubsection{Performance Drop using Context}
\label{subsubsec_error4}

\begin{figure}[t]
\centering
\includegraphics[width=0.6\linewidth]{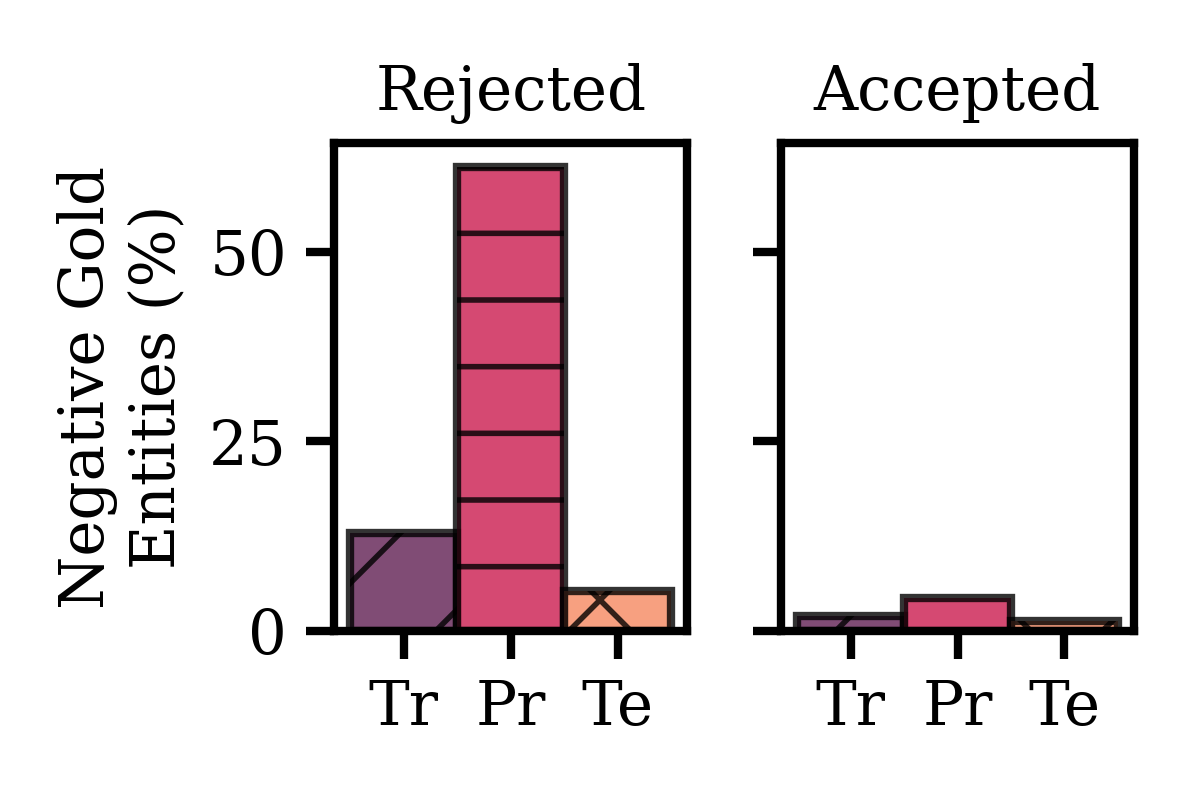}
\vskip -5pt
\caption{\small The majority of the rejected golden ``Problem'' entities are negative, leading to a performance drop in~\autoref{table_filter_context} when using context. Further discussion in~\autoref{subsubsec_error4}.}
\vskip -15pt
\label{fig_err_context}
\end{figure}

\autoref{subsubsec_result_filter_context} shows that there is a notable performance degradation for ``problem'' when context is provided. 
To investigate, we observe the ground-truth ``problem'' entities that are removed by the filter. Interestingly, we find that for most of them, the context specifically stated that the patient does not experience these problems. We then conduct further investigation on their \textit{polarity} attribute, which contains information on whether the patient is experiencing medical problems (or taking certain medications, for instance). To clarify, if there are explicit mentions that a patient does not have certain medical problems, the polarity is negative. Otherwise, it is positive. We conduct an analysis of how entity polarity affects filter responses. We plot our results in~\autoref{fig_err_context}.

\textbf{\textit{The ``negative'' polarity degrades performance}}. First, the dataset statistics in~\autoref{table_i2b2_dataset_stats} show that almost a fifth of ``problem'' entities are ``negative'', making it likely that these gold labels would be rejected. Furthermore,~\autoref{fig_err_context} reveals that the ``negative'' polarity causes the performance drop on the ``problem'' entity, as revealed by how a majority of the rejected gold ``problem'' entities are ``negative''.

\begin{table}[!t]
  \caption{i2b2 2012 Polarity Dataset Statistics}
  \label{table_i2b2_dataset_stats}
  \centering
  \footnotesize
  \begin{threeparttable}
      \begin{tabular}{
        @{\hspace{15pt}}c@{\hspace{12pt}}
        @{\hspace{12pt}}r@{\hspace{8pt}}
        @{\hspace{8pt}}r@{\hspace{8pt}}
        @{\hspace{8pt}}r@{\hspace{8pt}}
        @{\hspace{8pt}}r@{\hspace{15pt}}
      }
      \toprule

      \multirow{2}{*}{Polarity} & \multicolumn{4}{c}{\makecell{Entity Type (\# of samples)}} \\
      \cmidrule{2-5}
      & \texttt{Tr} & \texttt{Pr} & \texttt{Te} & \texttt{CD} \\
      \midrule
      Positive & 3684 & 4164 & 2544 & 996 \\
      Negative & 145 & 858 & 52 & 0 \\
      
      \bottomrule
      \end{tabular}
  \end{threeparttable}
  \vskip -11pt
\end{table}

\section{Conclusion and Future Work}

In this work, we propose a novel EDF framework to tackle the clinical named entity recognition task. Our comprehensive experiments demonstrate the strength of our framework across different dimensions. We also thoroughly investigate each framework component and provide several key insights. In future works, we hope to explore how to address the limitations of our work described in~\autoref{sec_limitation}.

\section{Limitations}
\label{sec_limitation}

First, we restrict our work to clinical narratives and have yet to explore how our framework generalizes to other texts. In this work, we deliberately focus on how well the method generalizes to different datasets, which (1) tackle different and clinically significant~\cite{lehman-etal-2022-learning} entity types, (2) are collected from different institutions (thus different distributions), (3) are de-identified in different ways (e.g., masks used for the patient’s sensitive information), (4) used different formats (e.g., header names and section organizations), etc. In fact, each patient is a unique case, and each of them can be treated as a separate domain~\cite{yang2023manydg}. Thus, generalizing to these datasets is already a significant challenge. However, testing how well our framework generalizes beyond clinical narratives would be an interesting avenue. Note that our motivation for this framework is that we found some clinical entities are easier to identify through simpler terms. This is particularly true for clinical narratives since most entities that are of interest~\cite{lehman-etal-2022-learning} follow this assumption. Thus, we designed our framework based on the characteristics of entities inside clinical narratives, not the narratives themselves. This is the reason we hypothesize that our framework may work outside clinical narratives (with similar entity characteristics). We leave this to future work.

Second, we restrict our work to only open-sourced models and leave experiments on proprietary models to future works. Most publicly available clinical narratives are under restrictive licenses. Hence, we cannot simply use commercial models, which may leak the data to a third party that is not covered by the restricted data use agreements. In contrast, open-sourced models have more practical values (e.g., they can be deployed). In this work, we deliberately use strong open-sourced models such as UniversalNER~\cite{zhou2023universalner}, which performs better than ChatGPT~\cite{brown2020language}. However, how open-sourced models fare with other proprietary models on clinical NER is still unknown. We leave them to future works.

Third, our work falls under the healthcare domain, a high-stakes setting. Despite the good performance of our methods, there is still a long way to reach the performance achieved in the original challenge of the datasets years ago, which are from the supervised learning settings. We provide their results in~\autoref{app_sup_ori}. It is also critical that NER performance in clinical texts can satisfy the high requirements set by healthcare applications. Nevertheless, our work paves a potential solution for the zero-shot clinical named entity recognition task.

Fourth, despite our attempts in generalization across datasets, it is very unfortunate that existing publicly available annotated clinical datasets only contain a few entity types. The issue is exacerbated by the significance of clinical domain, with emphasize on patient data protection (thus harder to publicly share) and the domain expertise to annotate. Once datasets with other entity types are made available, we will apply our methods to other entity types, which we are not able to do at this point given the lack of datasets with other entity types.

Fifth, our framework may provide higher inference cost than supervised methods. We remark that we develop a framework for a high-stakes domain that imposes high-performance requirements. Furthermore, our method is much more cost-efficient than curating an annotated datasets, which is required for supervised methods. Our framework is also flexible such that it provides trade-off between performance and inference cost depending on the user's needs.

\section{Ethic Statement}

Our research is conducted on open, retrospective clinical datasets without human subject intervention and thus will not harm human subjects. Furthermore, the clinical domain is complex and requires evaluation beyond performance, particularly regarding safety and bias. Unfortunately, the clinical narratives in our datasets are not associated with specific patients, impeding such evaluations. Further evaluations and validations from clinical experts will be needed to translate research into the clinical decision process.

\section{Acknowledgement}

Research reported in this publication was supported by the National Institute On Aging of the National Institutes of Health under Award Number R01AG082044. The content is solely the responsibility of the authors and does not necessarily represent the official views of the National Institutes of Health.

\bibliography{custom}

\newpage

\appendix
\label{appendix}

\section*{Appendix}
\label{app}

\section{Entity Decomposers}
\label{app_entity_decomposer}

We provide the details for each of our entity decomposition methods described in~\autoref{subsec_exp_decomposer} here:

\begin{itemize} [nosep,topsep=0pt,parsep=0pt,partopsep=0pt, leftmargin=*]
    \item \textbf{Manually curating a set of candidate types using expert-level knowledge}. Here, we refer to the annotation guidelines available in existing datasets since we believe they are curated by domain experts. For ``\texttt{Tr}'', ``\texttt{Pr}'', ``\texttt{Te}'' and ``\texttt{DD}'' we take the annotation guidelines from i2b2 2010. For ``\texttt{CD}'', we use i2b2 2012. For ``\texttt{AD}'' and ``\texttt{ADE}'', we use i2b2 2018 Task 2. We list the curated set in~\autoref{subsec_decomposer_annotation}.
    
    \item \textbf{Prompting an LLM for automatic generation}. We prompt ChatGPT with ``\textit{You are an intelligent clinical language model. Your job is to extract \{entity\_type\} from a patient's discharge summary. What entities can be considered as \{entity\_type\} in a discharge summary?}'' for each entity type. For reproducibility, we present the results in~\autoref{subsec_decomposer_chatgpt}.

    \item \textbf{Utilizing an existing medical knowledge bank}. We use the Unified Medical Language System (UMLS) since it contains standardized medical vocabulary for many clinical entities. Here, we take the UMLS semantic types for ``\texttt{Tr}'', ``\texttt{Pr}'' and ``\texttt{Te}'' available in i2b2 2010 guidelines. We list the curated set in~\autoref{subsec_decomposer_umls}.
\end{itemize}

\subsection{Annotation}
\label{subsec_decomposer_annotation}

\begin{flushleft}
    \texttt{\textbf{Treatment: } 
    medical treatment, medical intervention, medical procedure, medical device, treatment, biological substance, drug, medication
    }
\end{flushleft}
\begin{flushleft}
    \texttt{\textbf{Problem: } 
    medical problem, disease, syndrome, symptom, medical condition, behavior, virus, bacterium, injury, abnormality, abnormal test result, mental status
    }
\end{flushleft}
\begin{flushleft}
    \texttt{\textbf{Test: } 
    medical test, medical procedure, medical panel, medical examination, medical evaluation, test, procedure, laboratory procedure, diagnostic procedure, panel, measure, physiologic measure, vital sign, examination, evaluation
    }
\end{flushleft}
\begin{flushleft}
    \texttt{\textbf{Clinical Department: } 
    clinical department, medical department, clinical unit, clinical service, clinical practice, clinical room, department, location, building, hospital
    }
\end{flushleft}
\begin{flushleft}
    \texttt{\textbf{Disease/Disorder: } 
    medical problem, disease, syndrome, symptom, medical condition, behavior, virus, bacterium, injury, abnormality, abnormal test result
    }
\end{flushleft}
\begin{flushleft}
    \texttt{\textbf{Adverse Drug: } 
    drug
    }
\end{flushleft}
\begin{flushleft}
    \texttt{\textbf{Adverse Drug Event: } 
    medical problem
    }
\end{flushleft}

\begin{table}[!t]
  \caption{Dataset Statistics}
  \label{table_dataset_stats}
  \centering
  \begin{threeparttable}
      \begin{tabular}{
        @{\hspace{6pt}}l@{\hspace{6pt}}
        @{\hspace{6pt}}r@{\hspace{8pt}}
      }
      \toprule

      Dataset & \# samples \\
      \midrule
      i2b2 2010 & 27,625 \\
      i2b2 2012 & 7,446 \\
      i2b2 2018 & 9,181 \\
      CLEF 2014 & 10,422 \\
      
      \bottomrule
      \end{tabular}
  \end{threeparttable}
  \vskip -10pt
\end{table}
\begin{table*}[!t]
  \caption{Extension of~\autoref{table_main} for Precision ($\%$).}
  \label{table_precision}
  \centering
  \footnotesize
  \begin{threeparttable}
      \begin{tabular}{
        @{\hspace{2pt}}l@{\hspace{4pt}}
        @{\hspace{4pt}}l@{\hspace{4pt}}
        @{\hspace{4pt}}r@{\hspace{5pt}}
        @{\hspace{5pt}}r@{\hspace{5pt}}
        @{\hspace{5pt}}r@{\hspace{5pt}}
        @{\hspace{5pt}}r@{\hspace{5pt}}
        @{\hspace{4pt}}r@{\hspace{5pt}}
        @{\hspace{5pt}}r@{\hspace{5pt}}
        @{\hspace{5pt}}r@{\hspace{5pt}}
        @{\hspace{5pt}}r@{\hspace{5pt}}
        @{\hspace{5pt}}r@{\hspace{5pt}}
        @{\hspace{4pt}}r@{\hspace{3pt}}
        @{\hspace{6pt}}c@{\hspace{2pt}}
      }
      \toprule

      \multirow{2}{*}{Dataset} & \multirow{2}{*}{\makecell[l]{Entity\\Type}} & \multicolumn{5}{c}{UniNER} & \multicolumn{5}{c}{GNER} & \multirow{2}{*}{\makecell{UniNER-all\\(Supervised)}} \\
      \cmidrule(r){3-7}\cmidrule(r){8-12}
      && B & ED & F & EDF & $\Delta$ & B & ED & F & EDF & $\Delta$ & \\
      \midrule
      \multirow{3}{*}{i2b2 2010} & \texttt{Tr} & 51.63 & 30.35 & 61.71 & 55.09 & $+$3.46 & 46.08 & 27.19 & 74.16 & 66.60 & $+$20.52 & \textit{80.63} \\
      & \texttt{Pr} & 44.95 & 30.94 & 56.02 & 47.15 & $+$2.20 & 33.71 & 25.61 & 55.43 & 47.63 & $+$13.92 & \textit{75.87} \\
      & \texttt{Te} & 53.51 & 26.04 & 58.67 & 48.84 & $-$4.67 & 32.70 & 24.73 & 59.91 & 53.73 & $+$21.03 & \textit{79.14} \\
      \midrule
      \multirow{4}{*}{i2b2 2012} & \texttt{Tr} & 57.09 & 36.72 & 65.25 & 59.08 & $+$1.99 & 48.05 & 29.56 & 71.32 & 63.40 & $+$15.35 & \textit{81.10} \\
      & \texttt{Pr} & 42.93 & 32.65 & 53.03 & 46.87 & $+$3.94 & 37.49 & 28.71 & 52.72 & 46.57 & $+$9.08 & \textit{78.97} \\
      & \texttt{Te} & 51.35 & 21.58 & 58.34 & 47.04 & $-$4.31 & 29.96 & 18.15 & 57.95 & 49.69 & $+$19.73 & \textit{72.88} \\
      & \texttt{CD} & 35.87 & 19.94 & 54.85 & 47.56 & $+$11.69 & 55.11 & 11.73 & 57.17 & 50.77 & $-$4.34 & 59.19 \\
      \midrule
      CLEF 2014 & \texttt{DD} & 69.14 & 34.05 & 79.01 & 55.95 & $-$13.19 & 29.10 & 16.29 & 40.85 & 28.18 & $-$0.92 & \textit{78.71} \\
      \midrule
      \multirow{2}{*}{i2b2 2018} & \texttt{AD} & 12.43 & 4.54 & 20.93 & 15.09 & $+$2.66 & 1.67 & 2.43 & 6.20 & 8.79 & $+$7.12 & \textit{12.32} \\
      & \texttt{ADE} & 6.04 & 1.36 & 12.36 & 5.23 & $-$0.81 & 0.33 & 0.56 & 1.76 & 2.34 & $+$2.01 \\
      \midrule
      Avg. && 42.49 & 23.82 & 52.02 & 42.79 & $+$0.30 & 31.42 & 18.50 & 47.75 & 41.77 & $+$10.35 & \textit{64.76} \\
      \bottomrule
      \end{tabular}
  \end{threeparttable}
  \vskip -5pt
\end{table*}
\begin{table*}[!t]
  \caption{Extension of~\autoref{table_main} for Recall ($\%$).}
  \label{table_recall}
  \centering
  \footnotesize
  \begin{threeparttable}
      \begin{tabular}{
        @{\hspace{2pt}}l@{\hspace{4pt}}
        @{\hspace{4pt}}l@{\hspace{4pt}}
        @{\hspace{4pt}}r@{\hspace{5pt}}
        @{\hspace{5pt}}r@{\hspace{5pt}}
        @{\hspace{5pt}}r@{\hspace{5pt}}
        @{\hspace{5pt}}r@{\hspace{5pt}}
        @{\hspace{4pt}}r@{\hspace{5pt}}
        @{\hspace{5pt}}r@{\hspace{5pt}}
        @{\hspace{5pt}}r@{\hspace{5pt}}
        @{\hspace{5pt}}r@{\hspace{5pt}}
        @{\hspace{5pt}}r@{\hspace{5pt}}
        @{\hspace{4pt}}r@{\hspace{3pt}}
        @{\hspace{6pt}}c@{\hspace{2pt}}
      }
      \toprule

      \multirow{2}{*}{Dataset} & \multirow{2}{*}{\makecell[l]{Entity\\Type}} & \multicolumn{5}{c}{UniNER} & \multicolumn{5}{c}{GNER} & \multirow{2}{*}{\makecell{UniNER-all\\(Supervised)}} \\
      \cmidrule(r){3-7}\cmidrule(r){8-12}
      && B & ED & F & EDF & $\Delta$ & B & ED & F & EDF & $\Delta$ & \\
      \midrule
      \multirow{3}{*}{i2b2 2010} & \texttt{Tr} & 56.18 & 77.13 & 48.57 & 64.98 & $+$8.80 & 63.25 & 71.70 & 54.33 & 60.18 & $-$3.07 & \textit{70.02} \\
      & \texttt{Pr} & 55.56 & 65.28 & 49.87 & 56.64 & $+$1.08 & 50.39 & 60.12 & 46.83 & 54.03 & $+$3.64 & \textit{70.55} \\
      & \texttt{Te} & 44.81 & 63.84 & 31.02 & 40.32 & $-$4.49 & 43.32 & 68.94 & 27.01 & 39.88 & $-$3.44 & \textit{66.76} \\
      \midrule
      \multirow{4}{*}{i2b2 2012} & \texttt{Tr} & 52.11 & 70.34 & 44.37 & 57.45 & $+$5.34 & 52.95 & 62.61 & 45.61 & 51.92 & $-$1.03 & \textit{65.34} \\
      & \texttt{Pr} & 50.99 & 60.64 & 46.65 & 54.12 & $+$3.13 & 45.50 & 56.20 & 42.64 & 51.29 & $+$5.79 & \textit{71.70} \\
      & \texttt{Te} & 41.30 & 58.42 & 35.29 & 45.65 & $+$4.35 & 37.15 & 57.18 & 31.83 & 43.98 & $+$6.83 & 59.43 \\
      & \texttt{CD} & 49.04 & 88.78 & 25.18 & 32.56 & $-$16.48 & 63.20 & 79.88 & 29.83 & 30.03 & $-$33.17 & \textit{35.49} \\
      \midrule
      CLEF 2014 & \texttt{DD} & 35.29 & 70.62 & 31.71 & 60.75 & $+$25.46 & 13.10 & 29.99 & 11.90 & 25.74 & $+$12.64 & \textit{52.79} \\
      \midrule
      \multirow{2}{*}{i2b2 2018} & \texttt{AD} & 40.93 & 77.74 & 39.50 & 71.63 & $+$30.70 & 30.34 & 34.65 & 26.93 & 30.70 & $+$0.36 & \textit{17.24} \\
      & \texttt{ADE} & 22.86 & 52.88 & 22.27 & 48.11 & $+$25.25 & 3.78 & 16.90 & 3.78 & 15.31 & $+$11.53 & 34.00 \\
      \midrule
      Avg. && 44.91 & 68.57 & 37.44 & 53.22 & $+$8.31 & 40.30 & 53.82 & 32.07 & 40.31 & $+$0.01 & \textit{54.33} \\
      \bottomrule
      \end{tabular}
  \end{threeparttable}
  \vskip -5pt
\end{table*}

\subsection{ChatGPT}
\label{subsec_decomposer_chatgpt}

\begin{flushleft}
    \texttt{\textbf{Treatment: } 
    medical treatment, medication, medical procedure, therapy, medical intervention, consultation, counseling, discharge instruction, supportive care
    }
\end{flushleft}
\begin{flushleft}
    \texttt{\textbf{Problem: } 
    medical problem, medical diagnosis, disease, abnormal test result, symptom, abnormal imaging finding, complication, chronic health condition, medication side effect, mental health issue, social determinants of health
    }
\end{flushleft}
\begin{flushleft}
    \texttt{\textbf{Test: } 
    medical test, laboratory test, imaging study, diagnostic procedure, genetic test, electrodiagnostic test, functional test, microbiological test
    }
\end{flushleft}

\subsection{UMLS}
\label{subsec_decomposer_umls}

\begin{flushleft}
    \texttt{\textbf{Treatment: } 
    medical treatment, therapeutic procedure, preventive procedure, medical device, steroid, pharmacologic substance, biomedical material, dental material, antibiotic, clinical drug, drug delivery device
    }
\end{flushleft}
\begin{flushleft}
    \texttt{\textbf{Problem: } 
    medical problem, pathologic function, disease, syndrome, mental dysfunction, behavioral dysfunction, cell dysfunction, molecular dysfunction, congenital abnormality, acquired abnormality, injury, poisoning, anatomic abnormality, neoplastic process, virus, bacterium, symptom
    }
\end{flushleft}
\begin{flushleft}
    \texttt{\textbf{Test: } 
    medical test, laboratory procedure, diagnostic procedure
    }
\end{flushleft}

\section{Datasets}
\label{app_datasets}

We include all entities for i2b2 2010, ClinicalIE, and CLEF 2014.
For i2b2 2012, we found that UniversalNER and GNER performed poorly on the last two entities (e.g., evidence and occurrence) and decided to exclude them. We attribute this to them consisting mostly of verb phrases, while the training dataset consists mainly of noun entities. For i2b2 2018 Task 2, we test our method on a more challenging setup, extracting \textit{adverse drugs} and \textit{adverse drug events}~\cite{henry20202018}. We provide the dataset statistics in~\autoref{table_dataset_stats}.

\section{Recall and Precision Performance}
\label{app_recall_precision}

We provide the precisions and recalls for each dataset and entity type from~\autoref{table_main} in~\autoref{table_precision} and~\autoref{table_recall} respectively. We observe a similar trend for both metrics. Furthermore, we observe that UniNER benefits more from precision and GNER on recalls using our framework.

\section{Filter Prompt}
\label{app_filter_prompt}

We experiment with different ways to prompt in~\autoref{subsubsec_result_filter_prompt} and provide the specific instructions here.

\subsection{Without Description (Default)}
\begin{flushleft}
    \texttt{Can ’\textbf{\{entity\}}’ be considered a/an \textbf{\{entity\_type\}}? Answer with yes or no.
    }
\end{flushleft}

\subsection{With Description}
\begin{flushleft}
    \texttt{\textbf{Treatment: } 
    Can ’\textbf{\{entity\}}’ be considered a procedure or substance given to a patient to resolve a medical problem? Answer with yes or no.
    }
\end{flushleft}
\begin{flushleft}
    \texttt{\textbf{Problem: } 
    Can ’\textbf{\{entity\}}’ be considered an observation thought to be abnormal or caused by a disease? Answer with yes or no.
    }
\end{flushleft}
\begin{flushleft}
    \texttt{\textbf{Test: } 
    Can ’\textbf{\{entity\}}’ be considered a procedure or measure to find more information about a medical problem? Answer with yes or no.
    }
\end{flushleft}
\begin{flushleft}
    \texttt{\textbf{Clinical Department: } 
    Can ’\textbf{\{entity\}}’ be considered a clinical unit or clinical service name? Answer with yes or no.
    }
\end{flushleft}

\section{Few-shot Experiment}
\label{app_fewshot}

Here, we tried including some annotated samples in our framework and compared the approach to standard in-context learning. We randomly sample from the annotation guidelines and add them to the UniversalNER prompt. We also guarantee that there is at least one sample without entities of interest (e.g., sentence does not contain treatments or medical problems). Interestingly, we observe performance degradation across entity types the more samples we use. These contrastive results to general LLMs~\cite{xie2023empirical} further justify that open NER LLMs cannot be treated similarly to them. Furthermore, we observe that this also applies to our framework, although it is not as severe as standard in-context learning. 

We remark that performance drops on in-context learning are not uncommon. Previous works~\cite{zhao2021calibrate, zhu2023promptbench} show instability in performance for in-context learning. In addition, few-shot experiments are uncommon for zero-shot NER task~\cite{zhou2023universalner, ding2024rethinking, zaratiana2023gliner}, even if they use LLMs. Our work reveals that open NER LLMs may not benefit from in-context learning and are different from general LLMs. We leave further investigation to future works. 

\begin{figure}
\centering
\includegraphics[width=1\linewidth]{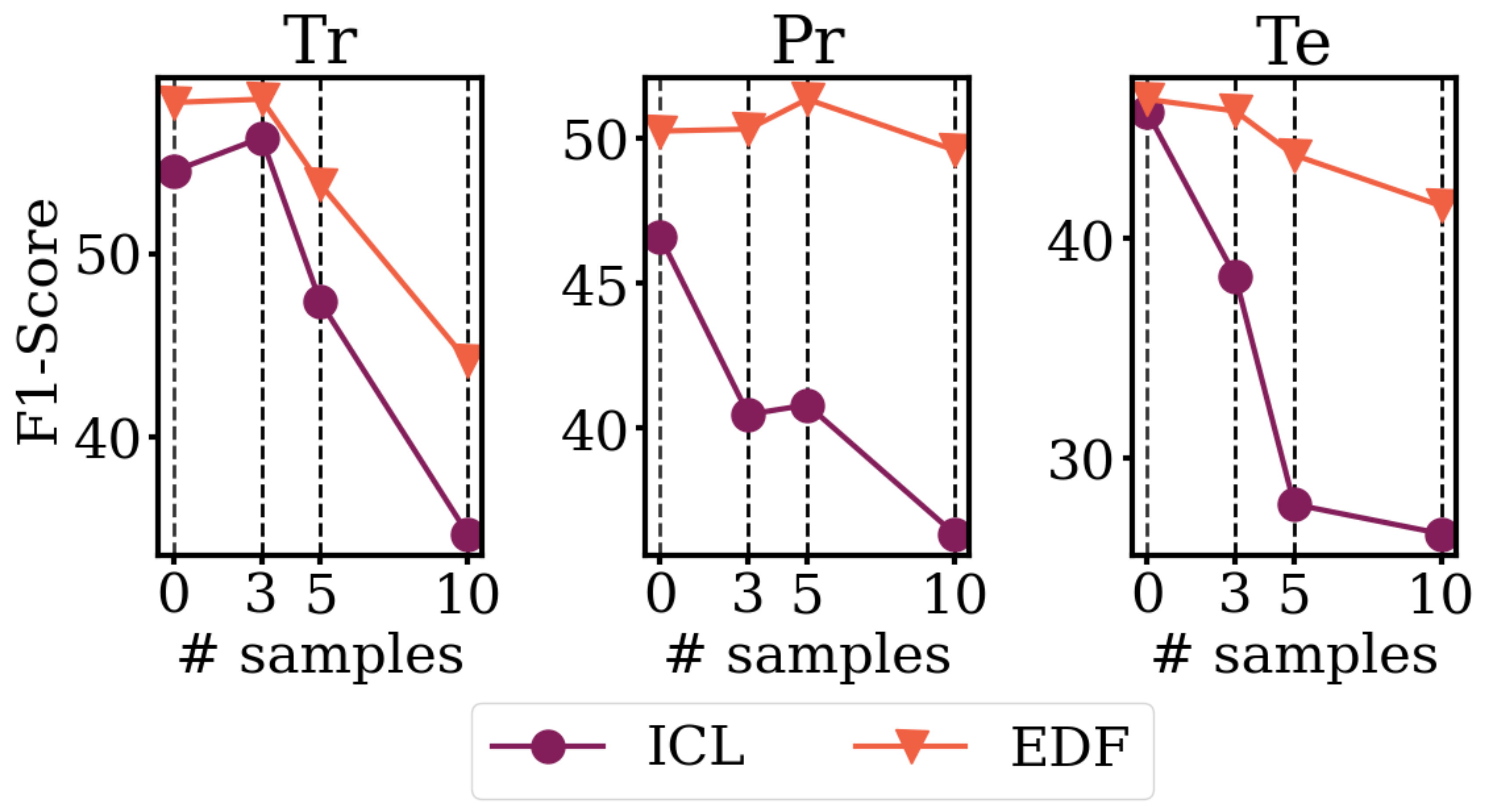}
\caption{\small \textbf{Few-shot performance comparison}. We observe performance drop using in-context learning (ICL). In contrast, our method (EDF) is more robust. We use the i2b2 2012 dataset with entity types treatment (\texttt{Tr}), problem (\texttt{Pr}), and test (\texttt{Te}).}
\vskip -10pt
\label{fig_fewshot}
\end{figure}

\begin{table}[!t]
  \caption{Performance on GLiNER.}
  \label{table_gliner}
  \centering
  \footnotesize
  \begin{threeparttable}
      \begin{tabular}{
        @{\hspace{0pt}}l@{\hspace{2pt}}
        @{\hspace{2pt}}l@{\hspace{2pt}}
        @{\hspace{2pt}}l@{\hspace{1pt}}
        @{\hspace{1pt}}r@{\hspace{2pt}}
        @{\hspace{2pt}}r@{\hspace{2pt}}
        @{\hspace{2pt}}r@{\hspace{2pt}}
        @{\hspace{2pt}}r@{\hspace{2pt}}
        @{\hspace{2pt}}r@{\hspace{0pt}}
      }
      \toprule

      \multirow{2}{*}{Dataset} & \multirow{2}{*}{\makecell[l]{Entity\\Type}} & \multirow{2}{*}{Metric} & \multicolumn{5}{c}{GLiNER} \\
      \cmidrule{4-8}
      &&& B & ED & F & EDF & $\Delta$ \\
      \midrule
      \multirow{9}{*}{i2b2 2010} & \multirow{3}{*}{\texttt{Tr}} & P & 52.03 & 35.70 & 70.79 & 66.71 & $+$14.68 \\
      && R & 44.55 & 76.13 & 39.86 & 63.93 & $+$19.38 \\
      \cmidrule{3-8}
      && F1 & 48.00 & 48.61 & 51.00 & 65.29 & $+$17.29 \\
      \cmidrule{2-8}
      & \multirow{3}{*}{\texttt{Pr}} & P & 71.19 & 48.48 & 79.13 & 67.32 & $-$3.87 \\
      && R & 49.22 & 63.36 & 46.16 & 56.49 & $+$7.27 \\
      \cmidrule{3-8}
      && F1 & 58.20 & 54.93 & 58.31 & 61.43 & $+$3.23 \\
      \cmidrule{2-8}
      & \multirow{3}{*}{\texttt{Te}} & P & 42.80 & 22.93 & 63.77 & 56.65 & $+$13.85 \\
      && R & 27.23 & 55.63 & 23.43 & 39.14 & $+$11.91 \\
      \cmidrule{3-8}
      && F1 & 33.28 & 32.47 & 34.27 & 46.30 & $+$13.02 \\
      \midrule
      \multirow{12}{*}{i2b2 2012} & \multirow{3}{*}{\texttt{Tr}} & P & 53.77 & 38.88 & 71.83 & 66.92 & $+$13.15 \\
      && R & 48.37 & 69.76 & 42.87 & 58.03 & $+$9.66 \\
      \cmidrule{3-8}
      && F1 & 50.93 & 49.93 & 53.69 & 62.16 & $+$11.23 \\
      \cmidrule{2-8}
      & \multirow{3}{*}{\texttt{Pr}} & P & 71.67 & 51.93 & 77.68 & 67.35 & $-$4.32 \\
      && R & 50.33 & 63.93 & 47.27 & 58.06 & $+$7.73 \\
      \cmidrule{3-8}
      && F1 & 59.13 & 57.32 & 58.78 & 62.36 & $+$3.23 \\
      \cmidrule{2-8}
      & \multirow{3}{*}{\texttt{Te}} & P & 43.97 & 19.72 & 66.72 & 55.42 & $+$11.45 \\
      && R & 39.17 & 60.09 & 35.17 & 48.41 & $+$9.24 \\
      \cmidrule{3-8}
      && F1 & 41.13 & 29.69 & 46.06 & 51.68 & $+$10.25 \\
      \cmidrule{2-8}
      & \multirow{3}{*}{\texttt{CD}} & P & 48.69 & 22.99 & 58.28 & 50.08 & $+$1.39 \\
      && R & 71.59 & 88.27 & 29.52 & 32.96 & $-$38.63 \\
      \cmidrule{3-8}
      && F1 & 57.96 & 36.48 & 39.19 & 39.76 & $-$18.20 \\
      \midrule  
      \multirow{3}{*}{CLEF 2014} & \multirow{3}{*}{\texttt{DD}} & P & 65.32 & 41.83 & 72.26 & 59.42 & $-$5.90 \\
      && R & 27.90 & 48.17 & 26.00 & 42.99 & $+$15.99 \\
      \cmidrule{3-8}
      && F1 & 39.09 & 44.78 & 38.24 & 49.89 & $+$10.80 \\      
      \midrule
      \multirow{6}{*}{i2b2 2018} & \multirow{3}{*}{\texttt{AD}} & P & 2.31 & 3.52 & 6.47 & 13.50 & $+$11.19 \\
      && R & 5.39 & 67.15 & 5.39 & 61.40 & $+$56.01 \\
      \cmidrule{3-8}
      && F1 & 3.23 & 6.69 & 5.88 & 22.13 & $+$18.90 \\   
      \cmidrule{2-8}
      & \multirow{3}{*}{\texttt{ADE}} & P & 7.42 & 2.17 & 15.03 & 7.70 & $+$0.28 \\
      && R & 14.31 & 44.93 & 13.12 & 40.95 & $+$26.64 \\
      \cmidrule{3-8}
      && F1 & 9.77 & 4.15 & 14.01 & 12.96 & $+$3.19 \\
      \midrule
      \multirow{3}{*}{Avg.} && P & 45.92 & 28.82 & 58.20 & 51.11 & $+$5.19 \\
      && R & 37.81 & 63.74 & 30.88 & 50.24 & $+$12.43 \\
      \cmidrule{3-8}
      && F1 & 40.10 & 36.51 & 39.94 & 47.40 & $+$7.29 \\        
      \bottomrule
      \end{tabular}
  \end{threeparttable}
  \vskip -5pt
\end{table}

\section{Performance on BERT-based Models}
\label{app_non_llm}

We use GLiNER~\cite{zaratiana2023gliner}, a BERT-based model for open-named entity recognition. Note that previous prompt engineering methods cannot be applied here. We conduct the experiment similar to UniNER and GNER, with Xie \emph{et.al.}~\cite{xie2023empirical} as our baseline. We present the results in~\autoref{table_gliner}. We observe the same trend as in~\autoref{subsec_overall_perf} with an average of $7.29\%$ F1-score improvement.

\section{Performance Drop on \texttt{CD}}
\label{app_drop_cd}

We observe significant recall drops to ``clinical department'' entities across all models. Here, we posit that some entities may not necessarily conform to the clinical department in the clinical domain. For instance, some entities are hospitals; thus, if a filter is prompted with our template (e.g., ``Can {hospitals} be considered as a clinical department?''), then it is likely to reject them. One possible solution is using a clear entity description. As illustrated in~\autoref{table_filter_ent_desc}, our framework outperforms the baseline (e.g. $45.97\%$ vs $38.66\%$ F1-score respectively) when using entity description.

\section{\Ours~Filter Precision/Recall Trade-off}
\label{app_filter_tradeoff}

\begin{figure}[!t]
\centering
\includegraphics[width=1\linewidth]{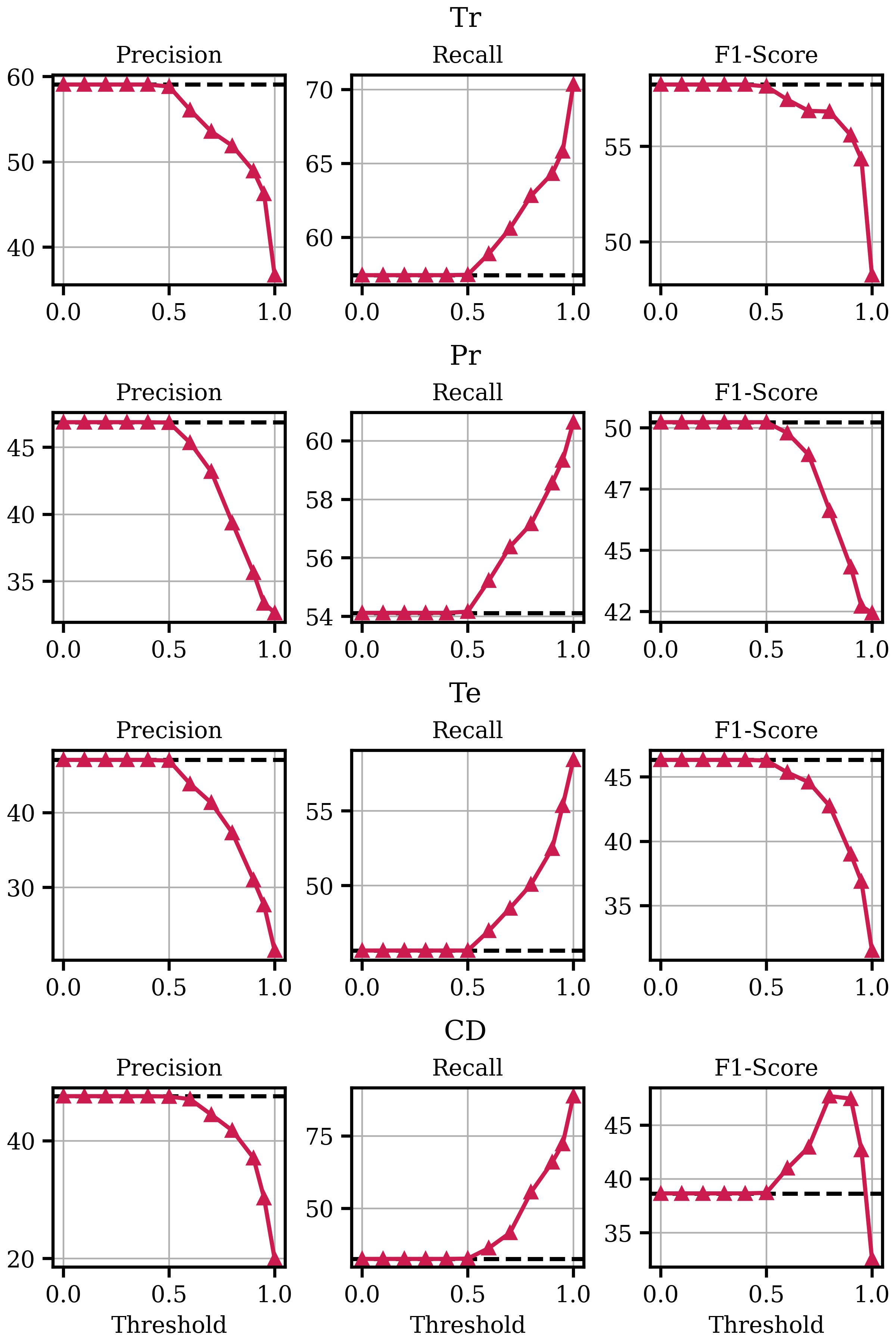}
\caption{\small Filter Precision/Recall Trade-off. There is an improvement in recall but a decrease in precision when increasing the threshold. The dashed line corresponds to performance with threshold $=0$. We use i2b2 2012 dataset.}
\vskip -15pt
\label{fig_filter_tradeoff}
\end{figure}

We analyze how filtering can be made more or less strict to achieve better trade-offs. We use the filter output probability to determine whether the entity is rejected or not. Concretely, rather than directly rejecting them if the filter outputs ``No'', we first look at the token probability. If it is less than a certain threshold, we then reject them. Our framework is simplified to entity decomposition if the threshold is $1$. We provide the results in~\autoref{fig_filter_tradeoff}. 

Overall, increasing the threshold leads to decreased precision and improved recall. Interestingly, better thresholding can improve the F1 Score in ``clinical department'' entities. This might be due to the noises for the entities as described in~\autoref{app_drop_cd}. 

\section{LLM Prompt Templates}

Our experiments involve large language models, which are often trained with specific templates. We use their default templates (except Llama2) throughout the experiments and present them here.

\subsection{UniNER}

\begin{flushleft}
    \texttt{A virtual assistant answers questions from a user based on the provided text.\\
    USER: Text: \textbf{\{input\}}\\
    ASSISTANT: I've read this text.\\
    USER: \textbf{\{instruction\}}\\
    ASSISTANT:
    }
\end{flushleft}

\subsection{GNER}

\begin{flushleft}
    \texttt{{[}INST{]} Please analyze the   sentence provided, identifying the type of entity for each word on a token-by-token basis.\\
    Output format is: word\_1(label\_1), word\_2(label\_2), ...\\
    We'll use the BIO-format to label the entities, where:\\
    1. B- (Begin) indicates the start of a named entity.\\
    2. I- (Inside) is used for words within a named entity but are not the   first word.\\
    3. O (Outside) denotes words that are not part of a named entity.\\
    \textbf{\{instruction\}}\\      
    Sentence: \textbf{\{input\}} {[}/INST{]}
    }
\end{flushleft}

\subsection{Asclepius}

\begin{flushleft}
    \texttt{You are an intelligent clinical   languge model.\\      
    Below is a snippet of patient's discharge summary and a following   instruction from healthcare professional.\\      
    Write a response that appropriately completes the instruction.\\      
    The response should provide the accurate answer to the instruction, while   being concise.\\
    \bigskip  
    {[}Discharge Summary Begin{]}\\      
    \textbf{\{input\}}\\      
    {[}Discharge Summary End{]}\\  
    \bigskip     
    {[}Instruction Begin{]}\\      
    \textbf{\{instruction\}}\\      
    {[}Instruction End{]}
    }
\end{flushleft}

\subsection{Llama2}

\begin{flushleft}
    \texttt{\textless{}s\textgreater{}{[}INST{]}   \textless{}\textless{}SYS\textgreater{}\textgreater\\      
    You are an intelligent clinical languge model.\\      
    Below is an instruction from healthcare professional.\\      
    Write a response that appropriately completes the instruction.\\      
    The response should provide the accurate answer to the instruction, while   being concise.\\      
    \textless{}\textless{}/SYS\textgreater{}\textgreater\\      
    \bigskip      
    \textbf{\{instruction\}} {[}/INST{]}
    }
\end{flushleft}

\section{LLM Hyperparameters}

We use the default hyperparameters for each model. For UniNER and GNER, we use greedy search. For Asclepius and Llama2, we use temperature $0.2$ and top $P$ probability $0.95$. Our exploration reveals consistent outputs for this set of hyperparameters.

\begin{table}[!t]
  \caption{Results from The Original Competitions}
  \label{table_sup_ori}
  \centering
  \begin{threeparttable}
      \begin{tabular}{
        @{\hspace{6pt}}l@{\hspace{6pt}}
        @{\hspace{8pt}}l@{\hspace{8pt}}
        @{\hspace{8pt}}r@{\hspace{8pt}}
      }
      \toprule

      Dataset & Metric & SOTA$^{\dagger}$ \\
      \midrule
      i2b2 2010 & Exact F1 & 85.20 \\
      i2b2 2012 & Span F Measure & 92.00 \\
      i2b2 2012 & Type Accuracy & 86.00 \\
      i2b2 2018 Task 2 & Lenient F1 & 58.73 \\
      CLEF 2014 & - & - \\
      
      \bottomrule
      \end{tabular}

      \begin{tablenotes}[normal,flushleft]
          \begin{footnotesize}
          \item[]
      $\dagger$ values reported from the best performing method in the challenges
          \par
          \end{footnotesize}
      \end{tablenotes}
  \end{threeparttable}
  \vskip -10pt
\end{table}

\section{Results from Original Challenges}
\label{app_sup_ori}

Each of the datasets in our experiments is curated from a competition~\cite{uzuner20112010, sun2013evaluating, henry20202018, mowery2014task}. We present the results from the original competitions in~\autoref{table_sup_ori}. The state-of-the-art methods from these competitions are trained using supervised learning. We remark that each competition is evaluated differently, and only i2b2 2010 use the same metric as ours.

\end{document}